# Multilingual Lexical Feature Analysis of Spoken Language for Predicting Major Depression Symptom Severity

Anastasiia Tokareva[1], Judith Dineley[1], Zoë Firth[1], Pauline Conde[1], Faith Matcham[2], Sara Siddi[3,4], Femke Lamers[5], Ewan Carr[1], Carolin Oetzmann[1], Daniel Leightley[6], Yuezhou Zhang[1], Amos A. Folarin[1,7], Josep Maria Haro[4], Brenda W.J.H. Penninx[5], Raquel Bailon[8,9], Srinivasan Vairavan[10], Til Wykes[1,7], Richard J.B. Dobson[1,11,12], Vaibhav A. Narayan[13], Matthew Hotopf[1,7], Nicholas Cummins[1,12], The RADAR-CNS Consortium[14]

1. Institute of Psychiatry, Psychology and Neuroscience, King's College London, London, UK
2. School of Psychology, University of Sussex, Falmer, UK
3. Impact and Prevention of Mental Disorders Research Group, Sant Joan de Déu Research Institut, Esplugues de Llobregat, Spain
4. Mental Health Networking Biomedical Research Centre (CIBERSAM), Madrid, Spain
5. Department of Psychiatry, Amsterdam Public Health Research Institute and Amsterdam Neuroscience, Amsterdam University Medical Centre, Vrije Universiteit , Amsterdam, NL
6. School of Life Course & Population Sciences, King's College London, London, England
7. South London and Maudsley NHS Foundation Trust, London, UK
8. Aragón Institute of Engineering Research (I3A), University of Zaragoza, Zaragoza, Spain
9. Biomedical Research Networking Centre in Bioengineering, Biomaterials and Nanomedicine (CIBER-BBN), Spain
10. Janssen Research and Development LLC, Titusville, NJ, United States
11. Institute of Health Informatics, University College London, London, UK
12. NIHR Biomedical Research Centre: Maudsley
13. Davos Alzheimer's Collaborative, Geneva, Switzerland
14. www.radar-cns.org

## Abstract

**Background**: Remotely captured spoken language could provide objective, regular indicators of depression symptom severity. However, research to date has largely used non-clinical, cross-sectional written language and complex machine learning (ML) approaches with limited interpretability.

**Methods:** We used linear mixed-effect models to identify interpretable lexical features associated with symptom severity in data from the RADAR-MDD study that comprised 5,846 smartphone recordings and Patient Health Questionnaire (PHQ-8) scores from 467 participants in the UK, Netherlands and Spain. We then developed ML models and systematically assessed via nested cross-validation whether interpretable lexical features or high-dimensional vector embeddings improved the accuracy of PHQ-8 prediction over sociodemographic and confounding features.

**Results:** Depression symptom severity was associated with five lexical features, including reductions in word count measures, use of first-person plural pronouns and positive word frequency. Associations were stable across countries, except for positive word frequency. Lexical features and vector embeddings did improve prediction accuracy beyond baseline models.

**Limitations:** Our cohort was skewed in age (median = 53, IQR 35 to 62) and majority female (n=357), potentially affecting the generalizability of our results. A lack of natural language processing tools for non-English languages restricted our feature choices.

**Conclusion:** Further research is required to realise the value of spoken lexical markers in clinical research and practice including larger and more diverse samples, elicitation protocol development and analytical methods that account for within- and between-individual variations.

## 1. Introduction

Major Depressive Disorder (MDD) affects over 10% of people worldwide (Bains & Abdijadid, 2023). It is associated with increased mortality risk (Chan et al., 2025) and a substantial economic burden across Europe (Pappa et al, 2024; Vieta et al., 2021; Wijnen et al., 2023). MDD often follows a chronic, relapsing-remitting course, with individuals experiencing recurrent episodes of depression interspersed with periods of partial or full remission. This pattern underscores the need for ongoing management strategies; however, a lack of scalable, remote measurement tools hinders the ability to monitor symptoms or identify early indicators of relapse (Kennis et al., 2020).

Remote Measurement Technologies (RMT), including smartphones or wearables, offer the potential of frequent, objective and non-invasive approaches for monitoring chronic conditions like MDD outside of clinical visits (Matcham et al., 2022; Andrews et al., 2023). Among the different physiological and behavioural signals capturable with RMT, such as sleep or physical activity, speech is a promising marker of MDD (Corbin et al., 2023). Depression severity has been associated with changes in both acoustic and linguistic patterns (Low et al., 2020; Koops et al., 2021). As RMT can potentially capture these changes repeatedly in daily life, speech offers a route to tracking symptom progression over time rather than only at isolated timepoints. This could enhance personalised medicine, as individual speech and language patterns could be tracked to assess symptom progression (Bickman et al., 2016).

To realise the potential of digital speech tools, it is important to understand how language characteristics, such as lexical features, change with MDD symptom severity. Lexical features refer to characteristics of words used in spoken language or text, such as word frequency and vocabulary diversity. They are often used to provide measurable insights into content and style of language (Khurana et al., 2022). Lexical features offer advantages for monitoring of MDD in that they offer quantifiable and interpretable markers of underlying cognitive and emotional processes (Trifu et al., 2024). For example, increases in the use of first-person pronouns and negative words have been associated with an increase in MDD severity (Savekar et al., 2022).

However, existing research on lexical features as digital markers of MDD has three main drawbacks. First, most studies have relied on cross-sectional, non-clinical datasets with limited sample sizes (Savekar et al., 2022). Results using such data seem promising but are likely to be underpowered (Ghasemzadeh et al., 2024). Second, many works investigate written rather than spoken language. For example, many studies have relied on written sources such as social media and online support forums, including Facebook and Reddit (Arachchige et al., 2021) and text messages (Tlachac et al., 2023). While benefitting from large-scale, readily available data, such resources typically lack clinical validation of participants' MDD status. Importantly, spoken language is also a different form of communication from written language in its mode of production, stylistic properties and disfluency patterns, and neurocognitive correlates (Huang et al., 2001; Yule, 2022). Written language can be edited and is more deliberate (Horowitz & Samuels, 1987), and we cannot presume findings from one form of expression will hold in the other. Finally, with the increasing popularity of large language models (LLMs) and transformers, many recent studies have focused on machine learning (ML) approaches applied to high-dimensional lexical representations rather than investigating specific characteristics (Low et al., 2020). While this shift toward high-dimensional representations can improve the predictive performances of models (Tahir et al., 2025), it comes at the expense of interpretability, making it difficult to understand which specific linguistic features are associated with depressive symptoms.

The first two of these limitations can be addressed through the choice of dataset. In this study, we utilise the Remote Assessment of Disease and Relapse in Major Depressive Disorder (RADAR-MDD) dataset (Matcham et al., 2022), which comprises repeated short speech samples, collected in four languages, taken from a longitudinal, observational study of a large clinical population of individuals with a history of MDD, supporting more robust and generalisable analyses than existing literature. Previous research has been undertaken on the RADAR-MDD speech dataset. Cummins et al. (2023) focused on assessing changes in acoustic and prosodic features across its three languages. Two studies employed ML approaches to predict MDD symptom severity using acoustic and prosodic information (White et al., 2025; Campbell et al., 2023). Two additional papers have explored language changes. Pérez-Toro et al. (2024) presented preliminary ML analyses utilising high-dimensional lexical representations. Zhang et al. (2024) used a deep learning model to identify depression-related topics in the English-language data. Both lexical analyses relied on high-dimensional representations and deep learning approaches with limited interpretability, reflecting the third limitation identified above.

Building on these works, the present study emphasises interpretability and explores the predictive value of lexical features for analysing MDD symptom severity. Our specific research objectives were to:

1. Explore whether specific interpretable lexical features, extracted from remotely collected spoken language samples, are robustly associated with MDD symptom severity.
2. Systematically investigate the impact of both input feature representation and ML model complexity on predictive performance compared to models trained using a baseline feature vector consisting of sociodemographic and confounding variables.

## 2. Methods

Our key methodological steps were collecting RADAR-MDD data, voice-to-text transcription, extracting both interpretable and high-dimensional lexical features, and analyzing the relationships between interpretable features and depression and their ability to predict depression severity compared to the more complex features (Figure 1). This approach addresses Objective 1, exploring associations between interpretable lexical features and MDD symptom severity, and Objective 2, a systematic comparison of the added value of lexical information in ML analyses. Each step is described in detail in the subsections below.

### 2.1 Study Design

RADAR-MDD was a longitudinal observational cohort study of individuals with a history of MDD from three recruitment sites: *King's College London* (KCL, London, United Kingdom); *Amsterdam UMC, Vrije Universiteit* (VUmc; Amsterdam, Netherlands); and *Centro de Investigación Biomédica en Red del Área Salud Mental* (CIBERSAM; Barcelona, Spain). Participants were recruited into the RADAR-MDD study between November 2017 and June 2020. The study enrolled 623 participants with a recent history of recurrent MDD, defined as an episode within the previous two years, and followed them for up to 24 months, with a final follow-up date of April 2021 (Matcham et al., 2022).

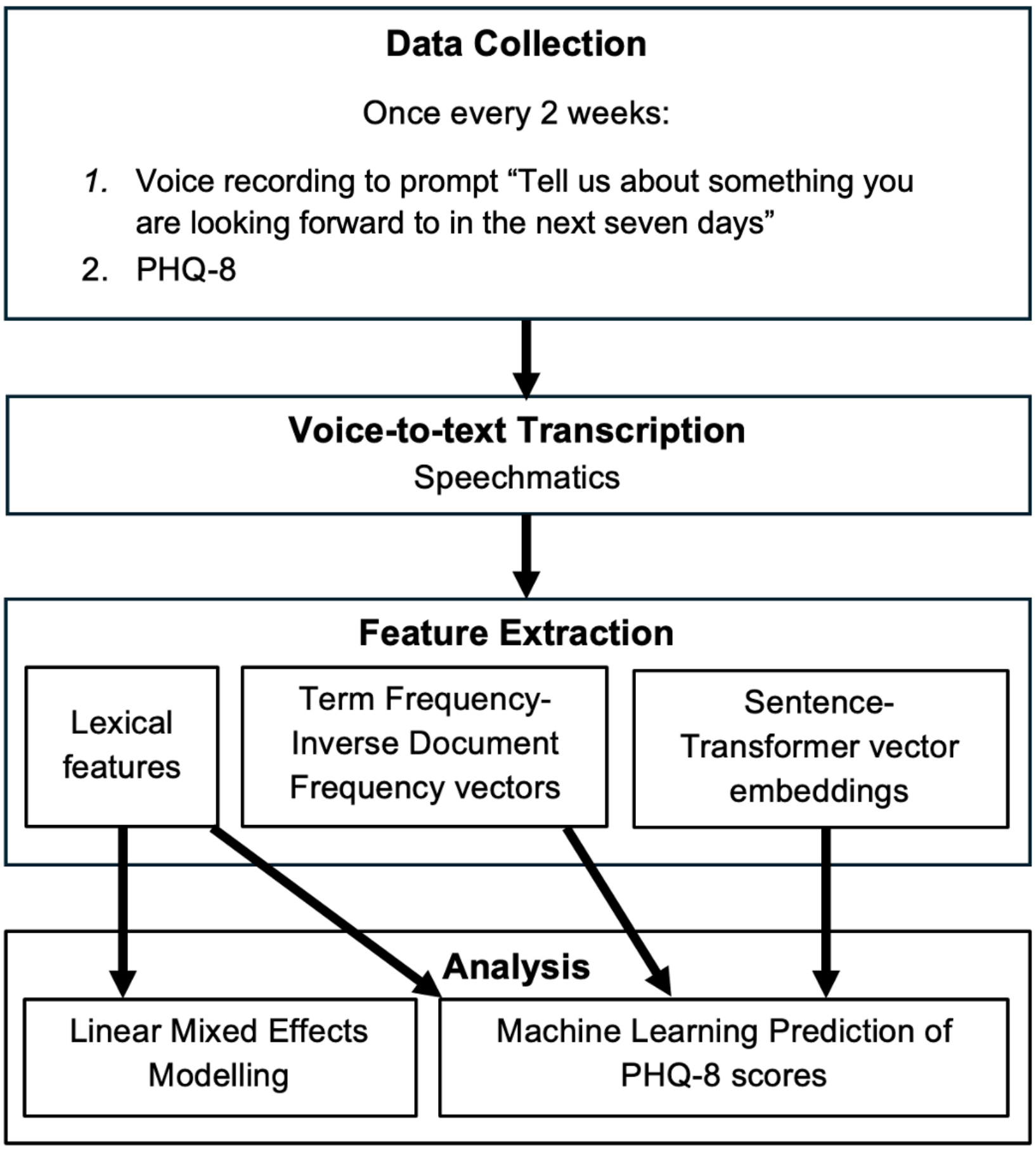

**Figure 1.** Overview of the key steps in the data processing pipeline: collection of the Remote Assessment of Disease and Relapse in Major Depressive Disorder (RADAR-MDD) data, automated voice-to-text transcription, extraction of interpretable and high-dimensional lexical features, and analysis of their relationships with depression symptom severity.

The main eligibility criteria included a non-psychotic MDD episode according to the Diagnostic and Statistical Manual of Mental Health Disorders (DSM-5) within the two years before enrolment and having recurrent MDD (lifetime history of at least two episodes). A history of MDD was confirmed for all participants before the start of data collection; see Matcham et al., (2019, 2022) for further details on the study protocol, inclusion, and exclusion criteria.

## 2.2 Ethics

Ethics approval for collection of RADAR-MDD data in London was granted by the Camberwell St Giles Research Ethics Committee (17/LO/1154), *Fundacio Sant Joan de Deu* Clinical Research Ethics Committee (CI: PIC-128-17) in Barcelona, and from the *Medische Ethische Toetsingscommissie VUmc* (2018.012–NL63557.029.17) in Amsterdam.

## 2.3 Patient Involvement

The RADAR-MDD protocol was co-developed with a patient advisory board that shared their views on several user-facing aspects of the study, including the choice and frequency of survey measures, the usability of the study app, participant-facing documents, selection of optimal participation incentives, selection, and deployment of wearable devices, as well as analysis plans. The speech task and subsequent analysis were discussed specifically with the RADAR-CNS Patient Advisory Board.

## 2.4. Speech Data Collection

Participants in the UK, Spain, and the Netherlands used an active Remote Measurement Technology (RMT) smartphone application developed for the study and installed on their Android devices (Ranjan et al., 2019). They were asked to complete speech tasks, speaking in the language native to their recruitment site (English, Dutch, Spanish or Catalan), once every two weeks, beginning in August 2019 and ending in April 2021. Participants were able to re-record their responses or skip the task entirely.

Speech tasks comprised a scripted task, where participants were asked to read aloud an excerpt of *The North Wind and The Sun*, and a free-response task with an open-ended cue "*Tell us about something you are looking forward to in the next seven days*" (Mundt et al., 2007). The scripted passage provided standardised linguistic content for acoustic analysis that was previously reported (Cummins et al., 2023) and was therefore not analysed in the present work. Instead, only responses to the open-ended task were analysed in this study, as they provided unscripted samples suitable for investigating lexical features.

To address safeguarding concerns and discourage participants from using the speech tasks to disclose suicidal thoughts or intentions, participants were informed that their free-speech recordings would not be monitored during data collection. Upon completion, the recordings were encrypted, linked to the participant's study ID, and securely transferred to a protected server.

Participation in the speech recordings was optional, and many participants contributed multiple times, allowing us to treat data collection as a repeated measures design. Participation levels are described in Cummins et al. (2023). Some of the recordings were obtained during COVID lockdowns in the UK, Netherlands and Spain, defined as the period from 23 March 2020 to 11 May 2021 in work by Leightley et al. (2021). Detailed information about the speech tasks can be found in Cummins et al (2023).

## 2.5. Outcome Measure

To assess the severity of MDD symptoms, the 8-item Patient Health Questionnaire (PHQ-8) scores (0-24) (Kroenke et al., 2009) were scheduled for completion once every two weeks concurrently with the speech recordings. Compared with the 9-item Patient Health Questionnaire (PHQ-9), the PHQ-8 used in our study omits the suicidality item, a choice reflecting the remote design that did not include real-time clinical monitoring or immediate intervention (Matcham et al., 2019). The PHQ-8 is an established and validated measure of MDD symptom severity (Kroenke et al., 2009).

## 2.6 Cohort Description and Bias Assessment

Sociodemographic variables, including age, sex at birth, and years of education (as a proxy for reading ability and verbal IQ), collected as part of the wider study protocol, have been shown to affect lexical aspects of speech (Le Dorze & Bedard, 1998). The variables were included as confounds in this analysis, consistent with the acoustic and prosodic analyses of the RADAR-MDD data (Cummins et al. 2023). Analyses of biases in the composition of analysable speech data in relation to sociodemographic and clinical variables were reported in the same work (Cummins et al., 2023).

## 2.7 Voice-to-Text Transcription

Recordings in English and Dutch were transcribed into text using the commercial ASR tool Speechmatics. Speechmatics is GDPR, SOC2 & HIPAA compliant; data security and privacy details are provided in their API documentation (Speechmatics, 2025). Speechmatics transcribes disfluencies, including repetitions and filler words (Russell et al., 2024), thereby enabling more accurate measurement of response length-dependent features such as word count and type-token ratio (Table 1), which were used in our analysis.

Recordings collected in Spain were spoken in either Spanish or Catalan and transcribed using Speechmatics' Spanish and Catalan ASR, respectively. Catalan transcripts were then translated into Spanish, also via Speechmatics software, enabling feature extraction via Spanish-language lexicons, to retain Catalan data in our sample for maximal statistical power. Speech data collected in Spain is referred to as Spanish in our results and discussion for conciseness.

We chose automatic transcription due to resource constraints that prevented time-consuming manual transcription of the large RADAR-MDD dataset. Similarly, we did not independently verify ASR or translation outputs due to feasibility and resource constraints; this limitation is covered in the Discussion.

## 2.8 Feature Extraction

To address Objective 1, we extracted a set of interpretable lexical features from the transcribed speech samples. We used Linguistic Inquiry and Word Count-22 (LIWC-22) software (Boyd et al., 2022) and custom Natural Language Toolkit (NLTK) 3.8.1 (Bird & Loper, 2004) code to extract the features from the transcriptions, separately for each language (English, Dutch, Spanish). Features were selected for analysis based on a thematic review of the literature of characteristics reported to be affected by MDD. A list of lexical features, LIWC extraction categories, word examples, and expected associations between features and depression symptom severity based on previous studies is provided (Table 1). As a starting point in our exploratory work, we sought to avoid multiple hypothesis testing, as this practice is associated with an increased risk of false positive findings (Ranganathan et al., 2016).

We used two complementary measures of response length, total word count (WC) and the number of words per sentence (WPS); and two measures of lexical diversity (LD), response-length-sensitive type-token ratio (TTR) and length-insensitive Brunet's index (Table 1).

To address Objective 2, we extracted high-dimensional text representations to enable systematic comparison with interpretable, feature-based approaches. Our more complex, high-dimensional features included term frequency-inverse document frequency (TF-IDF) vectors and dense vector embeddings. TF-IDF captures the importance of words in a text and is often used as a baseline for text-based ML prediction and comparison against more complex input like LLM embeddings (Jalilifard et al., 2021; Lin et al., 2023). TF-IDF vectors were obtained using the scikit-learn TfidfVectorizer (Pedregosa et al., 2011) with the number of features limited to the 500 most frequent terms across the dataset to reduce the risk of overfitting.

**Table 1.** Summary of lexical features and examples, LIWC categories, previous literature indicating an association between lexical features and MDD symptom severity and expected direction of change in features with increasing MDD symptom severity. The '↑' sign indicates an expected positive association between the feature and PHQ-8 scores, while '↓' indicates an expected negative association between the feature and PHQ-8 scores.

| Feature | Examples / Explanation (in English) | LIWC Category/ NLTK | Literature | Expected Change |
|---|---|---|---|---|
| Negative word frequency | "Hate", "worthless", "miserable" | Affect – emotion – emo_neg | Rude et al., 2004; Weintraub et al., 2023; Hur et al., 2024 | ↑ |
| Positive word frequency | "Happy", "pretty", "good" | Affect – emotion – emo_pos | Capecelatro et al., 2013; Himmelstein et al., 2018 | ↓ |
| Past-focus word frequency | "Went", "did", "been" | Time orientation - focuspast | Trifu et al., 2017, Spruit et al., 2022 | ↑ |
| Singular First-person pronoun frequency | "I", "myself", "mine" | Function – pronoun – ppron-i | Rude et al., 2004; Edwards & Holtzman, 2017; Zimmermann et al., 2016 | ↑ |
| Plural first-person pronoun frequency | "We", "us", "ours" | Function – pronoun – ppron-we | Ramirez-Esparza et al., 2008 | ↓ |
| Words per sentence | 5, 10, 15 | WPS – words per sentence | Trifu et al., 2017 | ↓ |
| Total word count | 15, 20, 25 | WC – word count | Trifu et al., 2017 | ↓ |
| TTR | Unique words / total number of words | NLTK | Botelho et al., 2024 | ↓ |
| Brunet's Index | Total text length ^ scaled power of unique words | NLTK | Botelho et al., 2024 | ↓ |

Transformer-based contextual embeddings are included in this systematic analysis as they are effective semantic representations for downstream prediction problems and are considered state-of-the-art across many Natural Language Processing (NLP) tasks (Tucudean et al., 2024). We extracted our embeddings using a sentence-transformer model, the paraphrase multilingual MPNet (masked and permuted pre-training for language understanding), available on Hugging Face (Reimers & Gurevych, 2019). This model was selected because it is optimised for all three languages of the RADAR-MDD transcripts (English, Dutch, Spanish) (Song et al., 2020). Base model version 2 was used, producing a 768-dimensional embedding.

Lexical features, continuous demographic variables including age and education years, and vector embeddings were pre-processed using standard scaling for normalization ($M = 0$, $SD = 1$). In addition, a COVID lockdown variable was dummy encoded as 1 if the recording was made during the lockdown and 0 if it was not. Categorical variables, sex and COVID lockdown status, were retained as raw features without scaling. TF-IDF was not scaled as the scores it produced were already normalized (Robertson, 2004).

## 2.9 Statistical Analysis

To address Objective 1, we used linear mixed effects (LME) models to estimate the relationships between lexical features and MDD symptom severity (PHQ-8 scores). The goal of this stage of the analysis was to determine whether specific interpretable features are robustly associated with depression severity while accounting for the repeated-measures and multiple collection sites in RADAR-MDD. This analysis comprised three steps. First, we pooled the data from each country and estimated associations across the combined dataset; second, we tested whether these associations differed by country and, third, provided country-specific effect estimates. We used LMEs, as they are well-suited for handling missing and repeated-measures data and can account for within-subject variability (Schober & Vetter, 2021).

### 2.9.1 Pooled Analysis

To facilitate comparison of effect sizes, lexical features and PHQ-8 score were standardised across the combined dataset prior to modelling, using the mean and standard deviation of the full dataset across all samples and countries. Each lexical feature was then modelled separately as the dependent variable. Each model included PHQ-8 score as the primary fixed effect of interest, adjusting for collection country, with age, sex, years of education, and COVID lockdown status included as confounding covariates. A random intercept for participant ID was included to account for inter-person variability in lexical features. All models were fitted in R using the *lme4* package (Bates et al., 2015). Models were estimated using restricted maximum likelihood (REML) to obtain unbiased estimates of variance components (Harville, 1977).

For each feature, estimated regression coefficients and 95% confidence intervals (CIs) were plotted using grouped forest plots in ggplot2 (Wickham, 2016). Relationships between lexical features and PHQ-8 scores were considered statistically significant where the 95% CI for the regression coefficient did not cross zero. Additionally, a heatmap of Pearson correlations between lexical features was created using *ggplot2* to aid interpretation of associations observed across multiple, potentially overlapping features.

### 2.9.2 Country Variation Analysis

To test whether lexical feature associations varied across countries, we fit an additional linear mixed-effects model that included a PHQ-8 × country interaction term. Likelihood ratio tests (LRT) were used to compare models with and without the interaction term; significance was assessed using chi-squared tests of the change in model deviance. Statistically significant interactions ($p<0.05$) were interpreted as evidence that the association between PHQ-8 scores and lexical features varied across countries, while non-significant results indicated broadly consistent slopes across countries. As likelihood ratio tests require models to be fitted on the same scale, these comparison models were estimated using maximum likelihood (ML) rather than REML (Pinheiro & Bates, 2000).

### 2.9.3 Country Stratified Analysis

Finally, we reran the analysis with a separate LME per country, retaining the same fixed and random effects structure as the primary analysis but excluding collection country, which was instead used to stratify the data into separate models. Grouped forest plots for features with a statistically significant country interaction are reported in the main text, with the remainder presented in the Supplementary Material.

## 2.10 Machine Learning Prediction

To address Objective 2, ML models to predict PHQ-8 scores were developed to assess the potential of interpretable lexical features and evaluate the added value of including the more complex TF-IDF vectors and MPNet vector embeddings. We conducted our ML analysis separately per country, as TF-IDF vectors are vocabulary dependent (Zhang et al., 2010), and vector embeddings can inherently capture language biases (Pérez-Toro et al., 2024, Cross et al., 2024, Caliskan et al., 2017).

To systematically evaluate the impact of increasing model and input complexity on predictive performance, we implemented a structured progression from baseline models trained on sociodemographic and confounding variables, to models incorporating interpretable lexical features, and finally to models using high-dimensional representations. This design allowed us to establish a realistic performance baseline and transparently assess whether high dimensional lexical representations offer additional predictive value.

We considered two regressor models, Elastic Net and XGBoost, as they have different approaches to handling bias-variance trade-off (i.e., balance between underfitting and overfitting (Briscoe & Feldman, 2011)), sensitivity to noise, and model complexity (Supplementary Table 1). Elastic Net was selected as a robust linear model that can handle multi-collinearity and feature sparsity through L1 and L2 regularization (Zhou and Hastie 2005). XGBoost was used as a non-linear model that could capture complex interactions between linguistic markers and PHQ-8 scores (Chen & Guestrin, 2016). Both models have been widely used for detecting depression using language (Fisher et al., 2026).

We also tested Support Vector Regressor (SVR) and Random Forest (RF) as alternate regression models. Additionally, we explored whether dimensionality reduction via truncated singular value decomposition (SVD, Falini, 2022), applied to the TF-IDF vectors (due to their sparsity) and for consistency to the MPNet embeddings, would improve predictive performance. Results from these analyses are presented in Supplementary Material.

Nested cross-validation was used to train and assess the models to provide a less biased estimate of model performance by separating model selection from model evaluation (Ghasemzadeh et al., 2024). An inner loop of five-fold cross-validation was used to optimize model performance (hyperparameters given in Supplementary Table 1), and an outer loop of five-fold cross-validation was used to assess generalizability on unseen data. All cross-validation folds were grouped by participant ID, so that all samples from a given participant were assigned to either a training or test set, enabling an accurate assessment of model generalizability. The models were evaluated using Root Mean Square Error (RMSE) and $R^2$ with RMSE used as the criterion for hyperparameter selection.

Each regressor was evaluated across four input configurations: (i) a baseline model consisting of age, sex, years of education, and COVID lockdown status only, noting these features were also included in all subsequent models; (ii) lexical features; (iii) lexical features and TF-IDF vectors; and (iv) MPNet embeddings. This design enabled a systematic comparison of whether gains in predictive performance justified increased representational and model complexity. Predictive performance values reported in the Results section are the averaged scores across the five outer loops. This analysis was undertaken in Python 3.10.9 using scikit-learn 1.4.2 (Pedregosa et al., 2011).

# 3. Results

## 3.1 Dataset description

We analyzed a total of 5,846 speech recordings from 467 individuals (Table 2). Most recordings were in English, with 57% of participants (N = 267) and 68% of recordings (N = 3,963). Most participants were female; women represented 78% of the English-speaking group, 70% of the Spanish sample, and 79% of the Dutch sample. Similarly, most recordings were made by female participants. Most recordings were made during COVID lockdown, comprising 68% of all English recordings (2,710/3,963), 78% of Spanish recordings (631/814), and 83% of Dutch recordings (877/1,059).

**Table 2:** A comparison of the sociodemographic, speech recordings and depression symptom severity (PHQ-8) distribution of the 467 participants who were enrolled in RADAR-MDD during the speech collection period.

| | | All (n = 467) | United Kingdom (n = 267) | The Netherlands (n = 98) | Spain (n = 102) |
|---|---|---|---|---|---|
| Sex assigned at birth | Female | 357 | 209 | 77 | 71 |
| | Male | 110 | 58 | 21 | 31 |
| Days of participation | Median | 585 | 702 | 459 | 667 |
| | IQR | 448-804 | 499-807 | 415-574 | 458-877 |
| Recordings per person | Median | 11 | 14 | 10 | 5 |
| | IQR | 4-19 | 6-21 | 4-17 | 2-12 |
| Recordings per female | Median | 10 | 13 | 13 | 8 |
| | IQR | 4-19 | 8-21 | 4-19 | 3-14 |
| Recordings per male | Median | 12 | 14 | 10 | 5 |
| | IQR | 4-18 | 6-21 | 4-16 | 2-11 |
| Recordings per person in lockdown | Median | 7 | 8 | 8 | 5 |
| | IQR | 2-14 | 3-16 | 2-15 | 1-9 |
| Age | Median | 53 | 53 | 52 | 53 |
| | IQR | 35-62 | 34-63 | 27-59 | 43-60 |
| Years in education | Median | 17 | 17 | 17 | 13 |
| | IQR | 13-19 | 13-19 | 14-21 | 10-16 |
| PHQ-8 score | Median | 8 | 8 | 8 | 11 |
| | IQR | 5-13 | 4-13 | 5-11 | 6-17 |

The median PHQ-8 score across recordings corresponded to mild depressive symptoms (median = 9, IQR 5 to 13). This was slightly higher in Spain (median = 11, IQR 6 to 17), consistent with moderate depressive symptoms. Plots for PHQ-8 distribution are provided in Supplementary Figure 1.

Additional information on the speech collection period and descriptive statistics for the entire RADAR-MDD dataset (including a scripted task not included in this investigation) has been reported previously (Cummins et al., 2023).

## 3.2 Mixed effects models

### 3.2.1 Pooled Results

Across the combined dataset, the mixed effects model revealed several statistically significant associations between PHQ-8 scores and lexical features, though standardised effect sizes were generally small to moderate in magnitude (Figure 2). PHQ-8 scores showed a positive association with TTR (Type Token Ratio) ($\beta$ = 0.067, 95% CI [0.030, 0.098]), and negative associations with words per sentence ($\beta$ = -0.091, 95% CI [-0.13, -0.055]), total word count ($\beta$ = -0.072, 95% CI [-0.10, -0.042]), positive word frequency ($\beta$ = -0.060, 95% CI -0.092, -0.027]), and first-person plural pronoun frequency ($\beta$ = -0.049, 95% CI [-0.13, -0.055]). A positive but not significant trend was observed for negative word frequency ($\beta$ = 0.026, 95% CI [-0.0014, 0.054]). Brunet's Index showed a negative but not significant trend ($\beta$ = -0.035, 95% CI [-0.069, 0.000]). No statistically significant associations were observed for past-focus word frequency or first-person singular pronoun frequency. The $\beta$ values and confidence intervals from this analysis are given in Supplementary Table 2.

To provide additional context for interpreting the results, we examined correlations among the lexical features by generating heatmaps that combine data from the three countries (Supplementary Figure 2). All demonstrate notable positive correlations (Pearson correlation coefficient >0.30) between total word count and Brunet Index, and a strong negative correlation (Pearson correlation coefficient <-0.30) between TTR and total word count.

### 3.2.2 Country Variation and Stratification Results

We found no evidence of differences in associations between linguistic features and PHQ-8 between the three countries (LRT $p > 0.05$), except for positive word frequency, which showed a significant country interaction ($\chi^2 = 10.63$, $p = .005$). This variation appears to be largely driven by a substantially stronger negative association in the Dutch cohort compared to the UK and Spanish cohorts (Figure 3). The $\chi^2$ values and p-values from this analysis are given in Supplementary Table 2, and the per-country forest plot and extracted β values and confidence intervals of the lexical features are given in Supplementary Figure 3 and Supplementary Table 3, respectively.

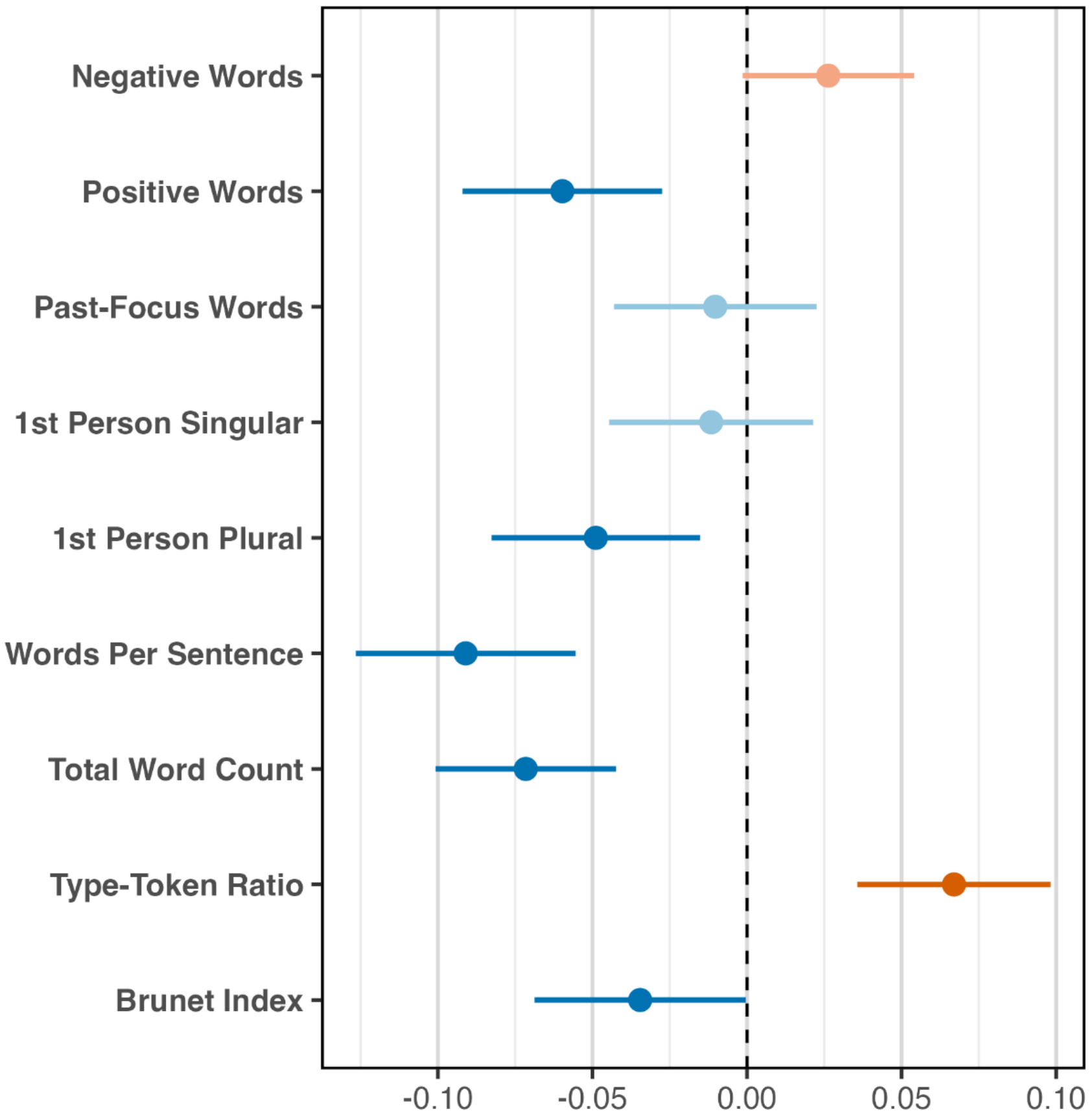

**Figure 2:** Fixed-effect estimates and 95% CIs from mixed effects modelling of the association between lexical features and PHQ-8 scores (n = 467 individuals; 5863 observations). Each point represents the standardised estimated effect of PHQ-8 scores on a lexical feature, with horizontal lines indicating 95% CIs. The vertical dashed black line represents the null effect (estimate = 0).

## 3.3. Machine learning prediction

The performance of our baseline models was close to chance, with RMSE values greater than five and $R^2$ values close to zero (Table 3). The addition of interpretable lexical features, TF-IDF vectors, or MPNet embeddings had very little influence on model performance. Given that PHQ-8 severity categories span 5-point bands, RMSE values of this magnitude mean that the average prediction error is large enough to shift a person into an adjacent severity category. The low $R^2$ values indicate that even our best-performing models are operating at a near-chance level, performing no better than simple mean-based prediction.

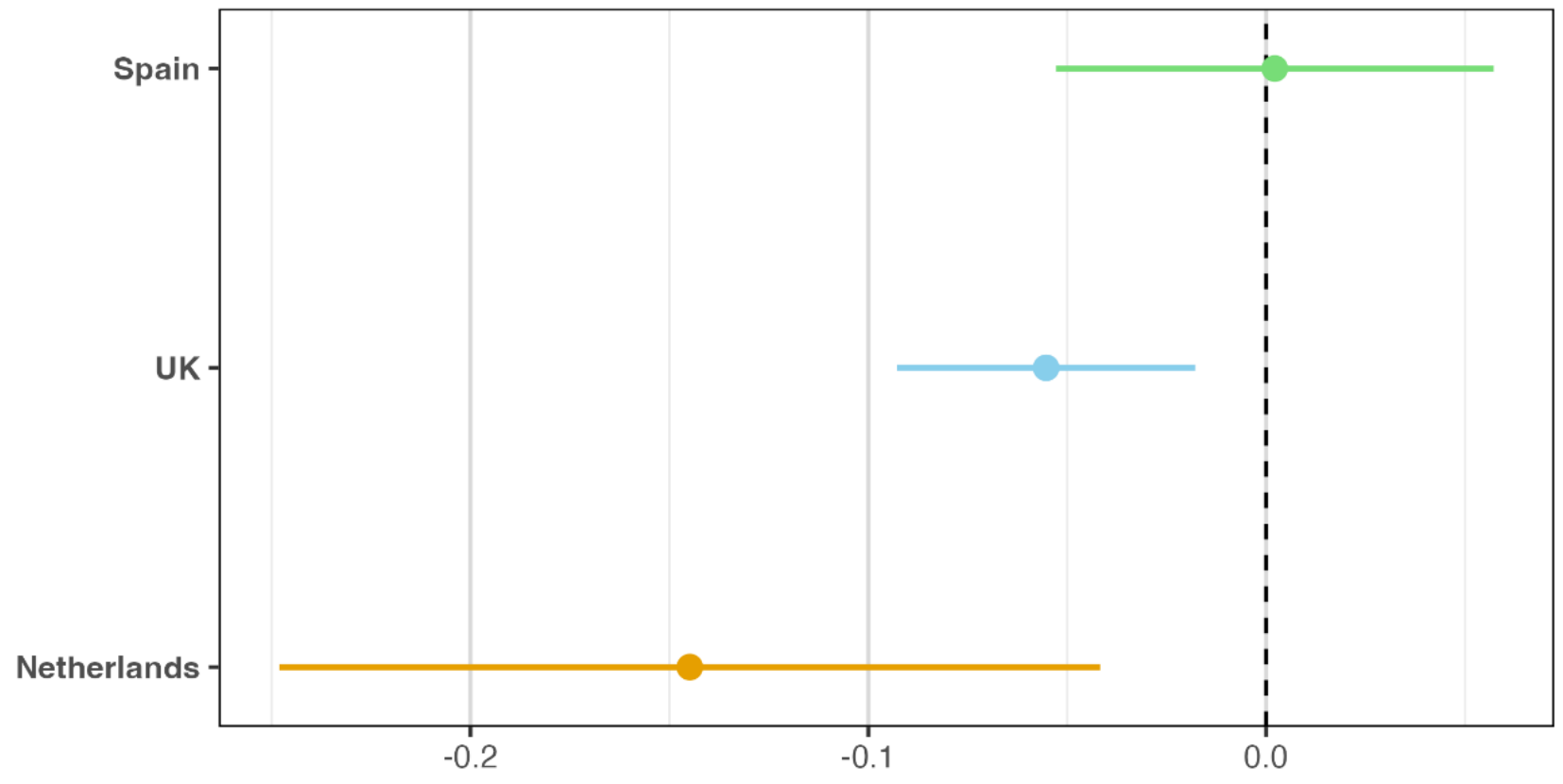

**Figure 3:** Per country fixed-effect estimates and 95% CIs from mixed effects modelling of the association between positive word use and PHQ-8 scores. Spain (n=102) is coloured green, the UK (n=267) is blue, and the Netherlands (n=98) is orange. Each point represents the standardised estimated effect of PHQ-8 scores on a lexical feature, with horizontal lines indicating 95% CIs. The vertical dashed black line represents the null effect (estimate = 0).

Additional results in the Supplementary Materials show that alternate regression models (SVR or RF) did not offer any meaningful improvement in predictive performance. Neither did applying truncated SVD to the TF-IDF vectors and MPNet embeddings. Complete ML results are provided in Supplementary Tables 4, 5, and 6.

## 4. Discussion

Using RADAR-MDD dataset, the largest clinical dataset of repeated speech recordings (Cummins et al., 2023), we report novel findings on the associations between depression symptom severity, as measured via remotely collected PHQ-8 questionnaires, and interpretable lexical features extracted from spoken responses to the prompt "*Tell us about something you are looking forward to in the next seven days*" (Objective 1). We also report findings on whether ML models can use lexical feature representations to predict symptom severity (Objective 2).

A strength of this work is the use of data collected from a clinical cohort as part of the RADAR-MDD multi-country longitudinal observational study. The RADAR-MDD dataset is unique in providing repeated recordings of natural speech collected using RMT from a clinical cohort recruited in the UK, Netherlands and Spain. In comparison, previous studies on lexical markers of MDD have used largely monolingual datasets drawn from written language resources, including social media and text messages (Arachchige et al., 2021; Tlachac et al., 2023).

**Table 3:** Summary of ML performance across different regressor models and languages. Each regressor was tested on four input configurations: (i) a baseline model using only age, sex, years of education, and COVID lockdown status, which were also carried into all subsequent models; (ii) lexical features alone; (iii) lexical features combined with TF-IDF vectors; and (iv) MPNet embeddings. Results in bold represent performance values for the highest-performing model, Elastic Net trained on MPNet embeddings.

| | | Language | | | | | |
|---|---|---|---|---|---|---|---|
| | | English (n = 267) | | Dutch (n = 98) | | Spanish (n = 102) | |
| **Regressor** | **Model** | **RMSE** | **$R^2$** | **RMSE** | **$R^2$** | **RMSE** | **$R^2$** |
| **Elastic Net** | Baseline | 5.74 | 0.01 | 5.04 | -0.02 | 6.53 | -0.02 |
| | All lexical | 5.70 | 0.02 | 4.93 | 0.02 | 6.53 | -0.02 |
| | All lexical + TF-IDF | 5.64 | 0.04 | 4.93 | 0.02 | 6.53 | -0.02 |
| | MPNet embeddings | **5.58** | **0.07** | **4.94** | **0.02** | **6.44** | **0.01** |
| **XGBoost** | Baseline | 5.79 | -0.00 | 5.01 | -0.01 | 6.91 | -0.13 |
| | All lexical | 5.80 | -0.01 | 5.01 | -0.01 | 6.82 | -0.10 |
| | All lexical + TF-IDF | 5.74 | 0.01 | 5.03 | -0.02 | 6.87 | -0.12 |
| | MPNet embeddings | 5.63 | 0.00 | 5.03 | -0.08 | 6.58 | -0.09 |

Our study resulted in two main findings. First, individuals with higher MDD symptom severity gave shorter, less lexically diverse responses, and used fewer positive words and first-person plural pronouns, although these associations were generally small to moderate in effect size. These associations were largely consistent across the three languages, with only positive word frequency showing cross-linguistic variation.

Second, our machine learning models could not reliably predict depression symptom severity, regardless of the lexical representation used. This suggests that previous reports of high ML performances for depression symptom severity using lexical feature representations may be overly optimistic.

## 4.1 Statistical Analysis of Lexical Features

Addressing Objective 1, using a pooled cross-language analysis, we observed that individuals with higher MDD symptom severity gave shorter, less diverse responses, and showed reduced use of first-person plural pronouns, regardless of the language spoken. Additionally, individuals with more severe symptoms used fewer positive words. However, effect sizes for these associations were generally small to moderate. LRT confirmed that these associations were largely consistent across the three languages, with positive word frequency being the only feature with significant cross-linguistic variation.

To the best of the authors' knowledge, this is the first time such multilingual associations have been reported for spoken language. Cross-cultural differences have previously been observed in the written language of individuals with depression (Loveys et al., 2018; De Choudhury et al., 2017). However, these works relied on social media data and are not directly comparable with spoken language samples collected under a consistent elicitation prompt.

Statistically significant negative associations were observed between PHQ-8 scores and both measures of response length (total word count and words per sentence), which were consistent with previously reported results from spoken narrative tasks (Trifu et al., 2017) and possibly reflected poverty of speech or lack of motivation in patients with depression (Cohen et al., 2014; Soini et al., 2024). The observed statistically significant positive association between the type-token ratio (TTR) and MDD symptom severity was unexpected, given prior literature (Botelho et al., 2024). However, TTR is sensitive to word count; with shorter text yielding higher TTR scores, creating an illusion of higher lexical diversity (Fergadiotis et al., 2015). Given the shorter response lengths observed among participants with more severe symptoms, this likely explains the unexpected direction. The Brunet's index, which is length-insensitive, showed a negative association in the expected direction (Botelho et al., 2024), although it was not significant. These differences highlight the benefit of using complementary measures when assessing the lexical diversity of speech (Lindsay et al., 2024).

Consistent with previous research, we found that individuals with greater depression symptom severity used more negative and fewer positive words (Hidalgo Julia et al., 2025), reflecting cognitive biases such as negative thinking (Rude et al., 2004).

However, as noted, positive word frequency was the only feature we observed showing significant cross-linguistic variation, with negative associations in our Dutch and English cohorts but not in our Spanish cohort. This result could be partially explained by Spanish exhibiting a higher bias towards positive word use than other languages (Dodds, Clark, Desu and Danforth, 2015). In addition, our analysis demonstrates that individuals with more severe depressive symptoms used fewer first-person plural pronouns in their speech, which follows the social disengagement model of depression (Ramirez-Esparza et al., 2008). Differing from prior research (Rude et al., 2004; Trifu et al., 2017), we did not observe any relationship between MDD severity and frequency of first-person singular pronouns or past-focus words.

We speculate that our prompt's anticipatory framing ("Tell us about something you are looking forward to in the next seven days") may have influenced several of our findings. It differs greatly from the retrospective, autobiographical response style text analysed in the literature. It is possible that the prompt did not encourage sustained reflection or longer responses if a participant could not immediately identify something to look forward to. This could have contributed to the shorter responses and higher TTR we observed in participants with higher depression symptomology. The same framing may have contributed to the absence of expected effects on first-person singular pronouns and past-focus words. The prompt may have encouraged participants to describe upcoming events rather than reflect on their current or past emotional experiences.

### 4.2 Machine Learning Prediction

Regarding Objective 2, we systematically investigated the impact of feature representation and ML model complexity on predictive performance. Our findings demonstrate that models that included lexical features did not yield meaningful gains in predictive performance over baseline models trained on sociodemographic and confounding variables.

In general, our ML models achieved near-chance-level predictive performance, regardless of the lexical representation used. These findings are novel in the context of spoken language and contrast sharply with the previously reported performances in the literature (Zhang et al., 2026). We speculate that this difference could be for two main reasons. First, we analyzed spoken language from a clinical population, compared to the use of written non-clinical samples in previous research (Arachchige et al., 2021; Tlachac et al., 2023). Second, we used a robust nested cross-validation approach, which provides a more reliable estimate of out-of-sample performance than alternative approaches (Ghasemzadeh et al., 2024).

Our results suggest that existing reports of predictive performance of lexical features in the MDD literature may be overly optimistic, reinforcing similar concerns in previous speech-based health research and highlighting the need for broader implementation of more robust cross-validation techniques (Berisha et al., 2022). The stronger performances observed within the inner loop of all regressors are consistent with this (Supplementary Tables 7-10), indicating that while models could learn from training data, they likely overfitted to a subgroup of participants and, therefore, struggled to generalize the patterns to the unseen test set (Aliferis & Simon, 2024). These findings reinforce the value of nested cross-validation for obtaining a realistic representation of performance. They may also reflect high inter-person variability in lexical markers of MDD (Tackman et al., 2019), making generalizable prediction using lexical features more challenging.

Interestingly, the use of contemporary language representations in our models did not demonstrably influence predictive performance. In contrast to other results in the literature (Yeskuatov et al., 2024), the combination of lexical features and TF-IDF vectors did not consistently improve the models we tested. Similarly, models using dense vector embeddings showed minimal improvements in predictive power compared to regressors trained exclusively on lexical features. These observations were unexpected, as vector embeddings capture richer semantic and contextual information compared to distinct lexical features (Wolfrath et al., 2024). In addition, the inclusion of the more complex input reduced interpretability of the models, limiting transparency of the underlying decision processes involved in predicting MDD severity (Stiglic et al., 2020).

Our lower predictive performance contrasts with the robust associations between MDD symptom severity and several lexical features identified through LME modelling. This discrepancy may arise because LMEs are better suited for repeated measures, as they account for within-person variability (Schober & Vetter, 2021). In contrast, the ML models we applied did not account for repeated measures or within-person variability; this might have limited our ability to capture person-specific patterns relevant to depression severity. Therefore, future work should investigate whether using models that account for intra-individual variability, such as mixed-effects or hierarchical approaches, can improve the predictive performance (Kilian et al., 2023; Hu et al., 2023; Hidalgo Julia et al., 2025, Plate et al., 2019).

### 4.3 Limitations

Several limitations of our project should be noted. Firstly, most participants in the RADAR-MDD dataset tended to be older (Matcham et al., 2022), and a higher proportion were women than in typical clinical MDD samples (Salk et al., 2017). This might have introduced bias, as sex and gender differences have been reported in lexical aspects of spoken language (Newman et al., 2008). Additionally, in RADAR-MDD, participants were allowed to re-record their responses or skip the task entirely, which may have influenced lexical content by allowing additional planning, editing, or self-monitoring. We did not record the number of re-record attempts made by participants, so we cannot offer analytical insights into whether this affected the results.

Secondly, a lack of NLP tools for Dutch and Spanish restricted our choice of analysable lexical features. For example, we would have liked to include absolutist language in our analysis; however, an absolutist lexicon was only available in English. This limitation highlights the ongoing imbalance in NLP resources and research, which remain disproportionately focused on English (Hovy & Prabhumoye, 2021). Related to this, our choice to translate Catalan into Spanish may have introduced noise into the lexical measures and embeddings.

Thirdly, our choice of sentence transformer was constrained to a multilingual model that supported all three languages. While this ensured compatibility, embedding quality might have been compromised by the "curse of multilinguality" (Conneau et al., 2020). Fourthly, dense vector embeddings that were used as ML input might have reflected gender and language biases present in the training data for Sentence-Transformers, such as a heavy focus on English datasets and text produced by male speakers (Hovy & Prabhumoye, 2021; Nemani et al., 2024).

Finally, ASR accuracy can vary widely between audio recordings (Kuhn et al., 2023). We did not evaluate the accuracy of the transcripts or translations in this dataset. To counter this, we chose a commercial transcription tool, which typically offers higher accuracy than open-source alternatives (Russell et al., 2024). The use of commercial systems poses cost and accessibility barriers for future replication, especially in resource-constrained settings (Russell et al., 2024). We did not perform a sensitivity analysis with an open-source ASR, as interpreting such results requires evaluating ASR output quality. This is a non-trivial task; standard metrics such as word error rate (WER) are too blunt to capture the relevance of individual errors to downstream lexical analyses, or their impact on model outcomes.

### 4.4 Concluding Remarks

Our analysis provides novel insights into the links between lexical features and depression severity across languages. Effect sizes in our statistical analysis were moderate, and the predictive performance of machine learning models using nested cross-fold validation was limited. These results point to generalisability challenges that warrant further investigation with larger, more diverse samples. The extracted features are specific to the protocol used, and our protocol design, including elicitation prompts, had no validated protocols to draw on (Cummins et al., 2024). Therefore, the value of remotely captured automatically extracted lexical markers in clinical research and practice cannot yet be fully understood from these initial exploratory findings.

Consequently, further research, informed by our experiences in RADAR-MDD, is needed, including studies that focus specifically on speech as the primary outcome measure and identifies prompts, or combinations of prompts, that elicit larger and more representative speech samples. Larger, multilingual speech datasets from diverse clinical populations are also required. An additional important question to investigate is whether observed differences in performance are driven by task framing (e.g., past- vs. future-oriented), modality (spoken vs. written), or their interaction. Addressing this will require studies that systematically vary these factors within the same population. It is also crucial to address the persisting bias towards English in NLP tool development and research to promote greater generalisability and inclusivity in speech and mental health research. From an ML perspective, it would be beneficial to explore the predictive power of alternative approaches, such as hierarchical or mixed-effect models, which can account for both within- and between-person variability (Rezende et al., 2022; Kilian et al., 2023). With such advances in protocols and analytical tools, the promise of spoken language measures integrated into personalised approaches to predict MDD symptom severity can be realised.

## CRedit authorship contribution statement

AT: Conceptualization, Methodology, Formal Analysis, Writing – Original Draft; JD: Conceptualization, Methodology, Formal Analysis, Data Curation, Writing – Original Draft, Supervision; ZF: Methodology, Writing – Review; PC: Data Curation, Software; FM, SS and FL: Data Curation, Project administration, Writing – Review; EC: Methodology, Writing – Review; CO: Data Curation, Writing – Review; DL: Data Curation, Project Administration, Writing – Review, YZ: Writing – Review; AF: Data Curation, Software, Project Administration; JMH: Project Administration, Funding Acquisition, Writing – Review; BP: Project administration, Writing – Review; RB and SV: Project administration, Writing – Review; TW: Project Administration, Funding Acquisition, Writing – Review; VN: Project Administration, Supervision, Funding Acquisition; MH: Methodology, Project Administration, Supervision, Funding Acquisition; NC: Conceptualization, Methodology, Validation, Data Curation, Writing – Original Draft, Supervision.

## Conflicts of Interest

RD is a director of CogStack Ltd and Onsentia Ltd. At the time this work was undertaken, NC was a consultant to thymia Limited and Noah Labs. All other authors have no other conflicts to declare.

## Acknowledgements

The RADAR-CNS project has received funding from the Innovative Medicines Initiative 2 Joint Undertaking under grant agreement No 115902. This Joint Undertaking receives support from the European Union's Horizon 2020 research and innovation programme and EFPIA (www.imi.europa.eu). We thank all the members of the RADAR-CNS patient advisory board for their contribution to the device selection procedures, and their invaluable advice throughout the study protocol design. This paper represents independent research part-funded by the National Institute for Health and Care Research (NIHR) Biomedical Research Centre (BRC): Maudsley. The views expressed are those of the authors and not necessarily those of the NIHR or the Department of Health and Social Care.

Participant recruitment in Amsterdam was partially accomplished through Hersenonderzoek.nl, a Dutch online registry that facilitates participant recruitment for neuroscience studies (https://hersenonderzoek.nl/). Hersenonderzoek.nl is funded by ZonMw-Memorabel (project no 73305095003), a project in the context of the Dutch Deltaplan Dementie, Gieskes-Strijbis Foundation, the Alzheimer's Society in theNetherlands and Brain Foundation Netherlands. Participants in Spain were recruited through the following in-stitutions: Parc Sanitari Sant Joan de D ́eu network of mental health services (Barcelona); Institut Catal'a de la Salut primary care services (Barcelona); Institut Pere Mata-Mental Health Care (Terrassa); Hospital Clínico San Carlos (Madrid). This research was reviewed by a team with experience of mental health problems and their carers who have been specially trained to advise on research proposals and documentation through the Feasibility and Acceptability Support Team for Researchers (FAST-R): a free, confidential service in England provided by the National Institute for Health Research Maudsley Biomedical Research Centre via King's College London and South London and Maudsley NHS Foundation Trust. We thank all GLAD Study volunteers for their participation, and gratefully acknowledge the NIHR BioResource, NIHR BioResource centres, NHS Trusts and staff for their contribution. We also acknowledge NIHR BRC, King's College London, South London and Maudsley NHS Trust and King's Health Partners. We thank the National Institute for Health Research, NHS Blood and Transplant, and Health Data Research UK as part of the Digital Innovation Hub Programme.

## Supplementary Material

**Supplementary Table 1:** Summary of hyperparameters that were optimized in the inner loop of each Machine Learning regressor model. Note that the results for the Support Vector Regressor and Random Forest are in the Supplementary Material.

| **Regressor** | **Hyper-parameter** | | | |
|---|---|---|---|---|
| Elastic Net (EN) | Alpha (0, 0.01, 0.1, 0.5, 1) | L1/L2 ratio (0.1, 0.5, 0.9) | | |
| EXtreme Gradient Boosting (XGBoost) | Learning rate (0.1, 0.01, 0.001) | Sub-sample (0.5, 0.7, 0.9) | L1 Alpha (0, 0.01, 0.1, 0.5, 1) | L2 Gamma (0, 0.01, 0.1, 0.5, 1) |
| Support Vector Regressor (SVR) | C (0.1, 0.5, 1, 5, 10), | Kernel (linear, radial basis function) | Gamma (scale, auto) | |
| Random Forest (RF) | Tree total number (50, 100, 200, 500) | Max tree depth (50, 100, 200, 500) | Min samples for node split (2, 5, 10) | |

**Supplementary Table 2:** Standardised β coefficients and 95% bootstrap CIs for linear mixed effects (LME) models that estimated the relationship between the 8-item Patient Health Questionnaire (PHQ-8) and each lexical feature, controlling for country, age, sex, years of education, and COVID lockdown status, with a random intercept per participant to account for repeated measures. Also reported are the p-values from the likelihood ratio test (LRT) comparing LME models with and without a PHQ-8 by country interaction term. Significant interactions are interpreted as evidence that the association between PHQ-8 scores and lexical features varied across countries, while non-significant results indicated broadly consistent slopes across languages.

| | ***β*** | **95% CI** | $\chi^2$ | **p-value** | **Interpretation** |
|---|---|---|---|---|---|
| Negative Word Frequency | 0.026 | [-0.002, 0.054] | 0.12 | 0.943 | Consistent across countries |
| Positive Word Frequency | -0.06 | [-0.092, -0.027] | 10.57 | 0.005 | Varies by country |
| Past-Focus Word Frequency | -0.01 | [-0.043, 0.023] | 3.81 | 0.149 | Consistent across countries |
| 1st Person Singular Pronouns | -0.012 | [-0.045, 0.021] | 2.33 | 0.312 | Consistent across countries |
| 1st Person Plural Pronouns | -0.049 | [-0.083, -0.015] | 0.77 | 0.681 | Consistent across countries |
| Words Per Sentence | -0.091 | [-0.127, -0.055] | 0.34 | 0.845 | Consistent across countries |
| Total Word Count | -0.072 | [-0.101, -0.043] | 1.56 | 0.458 | Consistent across countries |
| Type-Token Ratio | 0.067 | [0.036, 0.098] | 1.53 | 0.464 | Consistent across countries |
| Brunet's Index | -0.034 | [-0.069,0.000] | 4.96 | 0.084 | Consistent across countries |

**Supplementary Table 3:** Per-country standardised β coefficients and 95% bootstrap CIs for linear mixed effects (LME) models that estimated the relationship between the 8-item Patient Health Questionnaire (PHQ-8) and lexical features, controlling for age, sex, years of education, and COVID lockdown status, with a random intercept per participant to account for repeated measures.

| | **Dutch** | | **Spanish** | | **English** | |
|---|---|---|---|---|---|---|
| | ***β*** | **95% CI** | ***β*** | **95% CI** | ***β*** | **95% CI** |
| Negative Word Frequency | 0.014 | [-0.031, 0.060] | 0.047 | [-0.108, 0.201] | 0.021 | [0.006, 0.036] |
| Positive Word Frequency | -0.145 | [-0.248, -0.042] | 0.002 | [-0.053, 0.057] | -0.055 | [-0.093, -0.018] |
| Past-Focus Word Frequency | 0.072 | [-0.045, 0.189] | -0.042 | [-0.106, 0.022] | -0.016 | [-0.052, 0.019] |
| 1st Person Singular Pronouns | -0.041 | [-0.124, 0.041] | -0.029 | [-0.086, 0.028] | 0.000 | [-0.043, 0.042] |
| 1st Person Plural Pronouns | -0.053 | [-0.169, 0.062] | -0.017 | [-0.044, 0.011] | -0.054 | [-0.094, -0.013] |
| Words Per Sentence | -0.082 | [-0.165, 0.001] | -0.084 | [-0.182, 0.014] | -0.092 | [-0.135, -0.049] |
| Total Word Count | -0.046 | [-0.108, 0.015] | -0.046 | [-0.124, 0.033] | -0.080 | [-0.116, -0.044] |
| Type-Token Ratio | 0.057 | [-0.004, 0.118] | 0.022 | [-0.053, 0.096] | 0.077 | [0.036, 0.117] |
| Brunet's Index | -0.003 | [-0.044, 0.039] | 0.017 | [-0.065, 0.098] | -0.055 | [-0.101, -0.009] |

**Supplementary Table 4:** ML model performance across the inner and outer loops for the English dataset. Each regressor was tested on six input configurations: (i) a baseline model using only age, sex, years of education, and COVID lockdown status, which were carried into all subsequent models; (ii) lexical features alone; (iii) lexical features combined with TF-IDF vectors; (iv) truncated singular value decomposition (TSVD) dimensionality reduction applied to combined lexical and TF-IDF vectors; (v) MPNet embeddings; and (vi) TSVD dimensionality reduction applied to the MPNet embeddings. For both models utilizing TSVD, we retain the first 100 components. The results reported in the Results section are shaded.

| Regressor | Features | Inner Loop Performance | | Outer Loop Performance | |
|---|---|---|---|---|---|
| | | RMSE | $R^2$ | RMSE | $R^2$ |
| Elastic Net | Baseline | 5.64 | 0.06 | 5.74 | 0.01 |
| | Lexical | 5.58 | 0.08 | 5.70 | 0.02 |
| | Lexical + TF-IDF | 5.34 | 0.16 | 5.64 | 0.04 |
| | Lexical + TF-IDF, TSVD | 5.39 | 0.14 | 5.64 | 0.04 |
| | MPNet | 5.28 | 0.18 | 5.58 | 0.07 |
| | MPNet, TSVD | 5.46 | 0.12 | 5.68 | 0.03 |
| XGBoost | Baseline | 5.60 | 0.07 | 5.79 | -0.00 |
| | Lexical | 5.40 | 0.14 | 5.80 | -0.01 |
| | Lexical + TF-IDF | 5.28 | 0.17 | 5.74 | 0.01 |
| | Lexical + TF-IDF, TSVD | 4.10 | 0.50 | 5.69 | 0.03 |
| | MPNet | 4.13 | 0.49 | 5.63 | 0.00 |
| | MPNet, TSVD | 4.36 | 0.44 | 5.74 | -0.03 |
| Support Vector | Baseline | 5.63 | 0.06 | 5.90 | -0.05 |
| | Lexical | 5.46 | 0.12 | 5.77 | 0.00 |
| | Lexical + TF-IDF | 5.25 | 0.18 | 5.82 | -0.02 |
| | Lexical + TF-IDF, TSVD | 5.41 | 0.13 | 5.70 | 0.02 |
| | MPNet | 4.44 | 0.38 | 5.72 | 0.02 |
| | MPNet, TSVD | 5.02 | 0.25 | 5.71 | 0.02 |
| Random Forest | Baseline | 3.50 | 0.64 | 6.82 | -0.43 |
| | Lexical | 3.09 | 0.71 | 5.70 | 0.02 |
| | Lexical + TF-IDF | 3.16 | 0.70 | 5.97 | -0.08 |
| | Lexical + TF-IDF, TSVD | 2.17 | 0.86 | 5.66 | 0.04 |
| | MPNET | 1.95 | 0.89 | 5.67 | -0.01 |
| | MPNET, TSVD | 2.28 | 0.88 | 6.56 | -0.07 |

**Supplementary Table 5:** ML model performance across inner and outer loop for the Dutch dataset. Each regressor was tested on six input configurations: (i) a baseline model using only age, sex, years of education, and COVID lockdown status, which were carried into all subsequent models; (ii) lexical features alone; (iii) lexical features combined with TF-IDF vectors; (iv) truncated singular value decomposition (TSVD) dimensionality reduction applied to combined lexical and TF-IDF vectors; (v) MPNet embeddings; and (vi) TSVD dimensionality reduction applied to the MPNet embeddings. For both models utilizing TSVD, we retain the first 100 components. The results reported in the Results section are shaded.

| Regressor | Features | Inner Loop Performance | | Outer Loop Performance | |
|---|---|---|---|---|---|
| | | RMSE | $R^2$ | RMSE | $R^2$ |
| Elastic Net | Baseline | 4.90 | 0.05 | 5.04 | -0.02 |
| | Lexical | 4.84 | 0.07 | 4.93 | 0.02 |
| | Lexical + TF-IDF | 4.84 | 0.07 | 4.93 | 0.02 |
| | Lexical + TF-IDF, TSVD | 4.81 | 0.08 | 4.95 | 0.02 |
| | MPNet | 4.65 | 0.14 | 4.94 | 0.02 |
| | MPNet, TSVD | 5.81 | 0.22 | 6.68 | -0.06 |
| XGBoost | Baseline | 5.60 | 0.07 | 5.79 | -0.00 |
| | Lexical | 4.91 | 0.04 | 5.01 | -0.01 |
| | Lexical + TF-IDF | 4.62 | 0.15 | 5.03 | -0.02 |
| | Lexical + TF-IDF, TSVD | 3.82 | 0.42 | 4.98 | 0.00 |
| | MPNet | 3.65 | 0.46 | 5.03 | -0.08 |
| | MPNet, TSVD | 3.71 | 0.45 | 5.07 | -0.10 |
| Support Vector | Baseline | 4.78 | 0.09 | 5.33 | -0.16 |
| | Lexical | 4.78 | 0.09 | 5.01 | -0.01 |
| | Lexical + TF-IDF | 4.77 | 0.10 | 5.03 | -0.02 |
| | Lexical + TF-IDF, TSVD | 4.78 | 0.09 | 5.02 | -0.02 |
| | MPNet | 4.61 | 0.15 | 5.06 | -0.03 |
| | MPNet, TSVD | 4.63 | 0.15 | 5.08 | -0.04 |
| Random Forest | Baseline | 3.34 | 0.56 | 5.94 | -0.45 |
| | Lexical | 2.38 | 0.77 | 5.57 | -0.25 |
| | Lexical + TF-IDF | 2.96 | 0.65 | 5.31 | -0.13 |
| | Lexical + TF-IDF, TSVD | 1.94 | 0.85 | 5.03 | -0.02 |
| | MPNET | 1.81 | 0.87 | 5.14 | -0.12 |
| | MPNET, TSVD | 2.16 | 0.81 | 5.11 | -0.11 |

**Supplementary Table 6:** ML model performance across inner and outer loop for the Spanish dataset. Each regressor was tested on six input configurations: (i) a baseline model using only age, sex, years of education, and COVID lockdown status, which were carried into all subsequent models; (ii) lexical features alone; (iii) lexical features combined with TF-IDF vectors; (iv) truncated singular value decomposition (TSVD) dimensionality reduction applied to combined lexical and TF-IDF vectors; (v) MPNet embeddings; and (vi) TSVD dimensionality reduction applied to the MPNet embeddings. For both models utilising TSVD, we retain the first 100 components. The results reported in the Results section are shaded.

| Regressor | Features | Inner Loop Performance | | Outer Loop Performance | |
|---|---|---|---|---|---|
| | | RMSE | $R^2$ | RMSE | $R^2$ |
| Elastic Net | Baseline | 6.22 | 0.11 | 6.53 | -0.02 |
| | Lexical | 6.19 | 0.12 | 6.53 | -0.02 |
| | Lexical + TF-IDF | 6.19 | 0.12 | 6.53 | -0.02 |
| | Lexical + TF-IDF, TSVD | 6.21 | 0.11 | 6.64 | -0.05 |
| | MPNet | 5.71 | 0.25 | 6.44 | 0.01 |
| | MPNet, TSVD | 5.81 | 0.22 | 6.68 | -0.06 |
| XGBoost | Baseline | 5.23 | 0.35 | 6.91 | -0.13 |
| | Lexical | 5.17 | 0.36 | 6.82 | -0.10 |
| | Lexical + TF-IDF | 5.24 | 0.33 | 6.87 | -0.12 |
| | Lexical + TF-IDF, TSVD | 4.90 | 0.05 | 5.04 | -0.02 |
| | MPNet | 3.45 | 0.71 | 6.58 | -0.09 |
| | MPNet, TSVD | 3.25 | 0.71 | 6.61 | -0.09 |
| Support Vector | Baseline | 6.16 | 0.13 | 6.76 | -0.09 |
| | Lexical | 6.06 | 0.16 | 6.65 | -0.06 |
| | Lexical + TF-IDF | 6.02 | 0.17 | 6.76 | -0.09 |
| | Lexical + TF-IDF, TSVD | 5.99 | 0.17 | 6.74 | -0.08 |
| | MPNet | 5.43 | 0.33 | 6.42 | 0.02 |
| | MPNet, TSVD | 5.42 | 0.33 | 6.41 | 0.02 |
| Random Forest | Baseline | 3.09 | 0.78 | 7.78 | -0.44 |
| | Lexical | 2.25 | 0.88 | 7.55 | -0.37 |
| | Lexical + TF-IDF | 2.05 | 0.90 | 7.32 | -0.28 |
| | Lexical + TF-IDF, TSVD | 2.48 | 0.86 | 6.64 | -0.05 |
| | MPNET | 2.14 | 0.89 | 6.51 | -0.07 |
| | MPNET, TSVD | 2.55 | 0.85 | 6.56 | -0.07 |

**Supplementary Figure 1:** A comparison of the distribution of *8-item Patient Health Questionnaire* (PHQ-8) scores for the UK (left; n = 3,963), Dutch (middle; n = 1,059) and Spanish (n = 814) participants who provided analysable speech data.

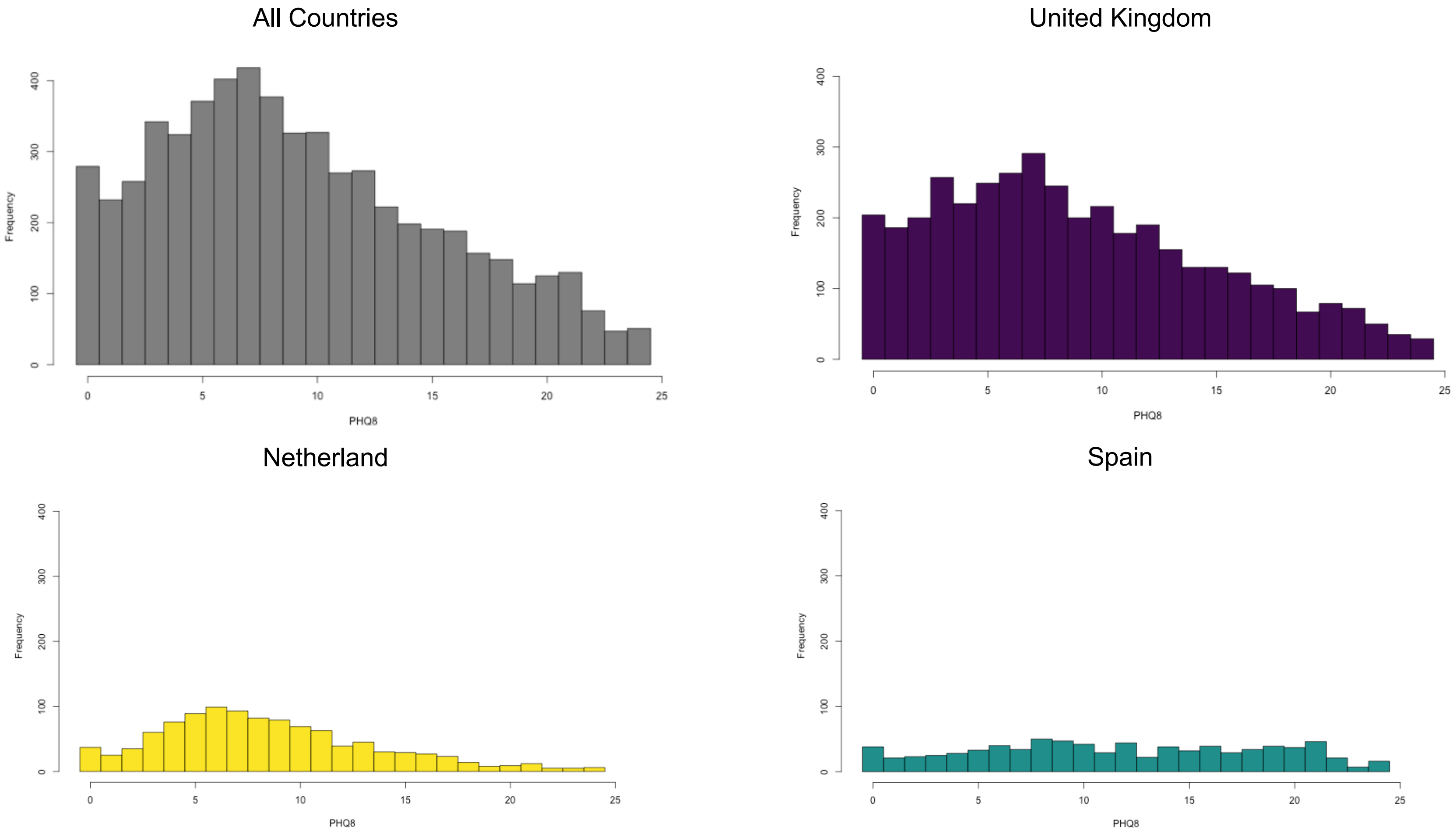

**Supplementary Figure 2:** Heatmap of Pearson correlation coefficients between interpretable lexical features, combining data from the three countries.

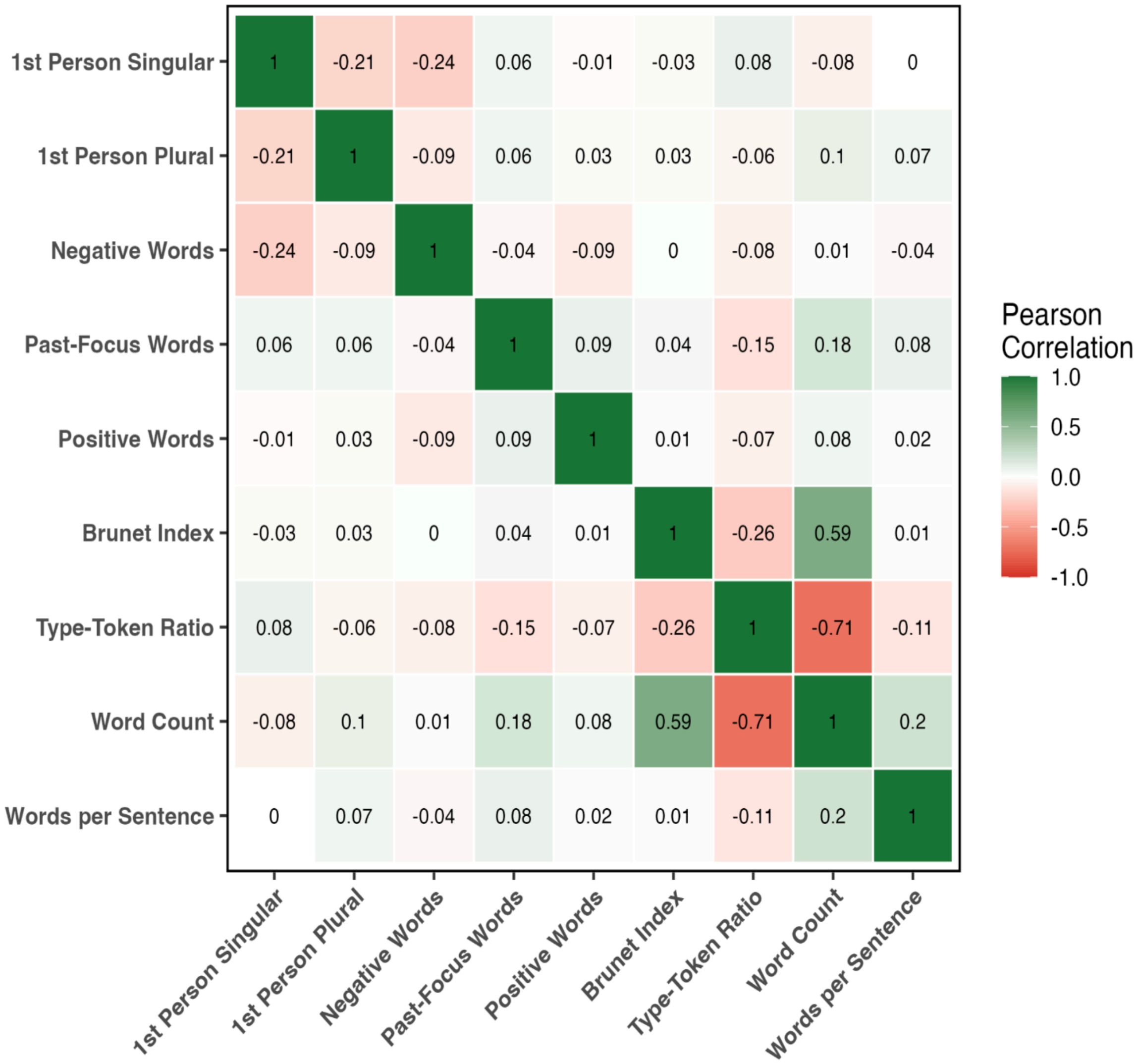

**Supplementary Figure 3:** Association of speech features with PHQ-8 scores across the three languages (n = 467 individuals; 5863 observations). Notes. This figure shows fixed-effect estimates and 95% CIs from mixed-effects modelling of the association between lexical features and PHQ-8 scores. Each point represents the standardised estimated effect of PHQ-8 scores on a lexical feature, with horizontal lines indicating 95% CIs. The vertical dashed black lines represent the null effect (estimate = 0).

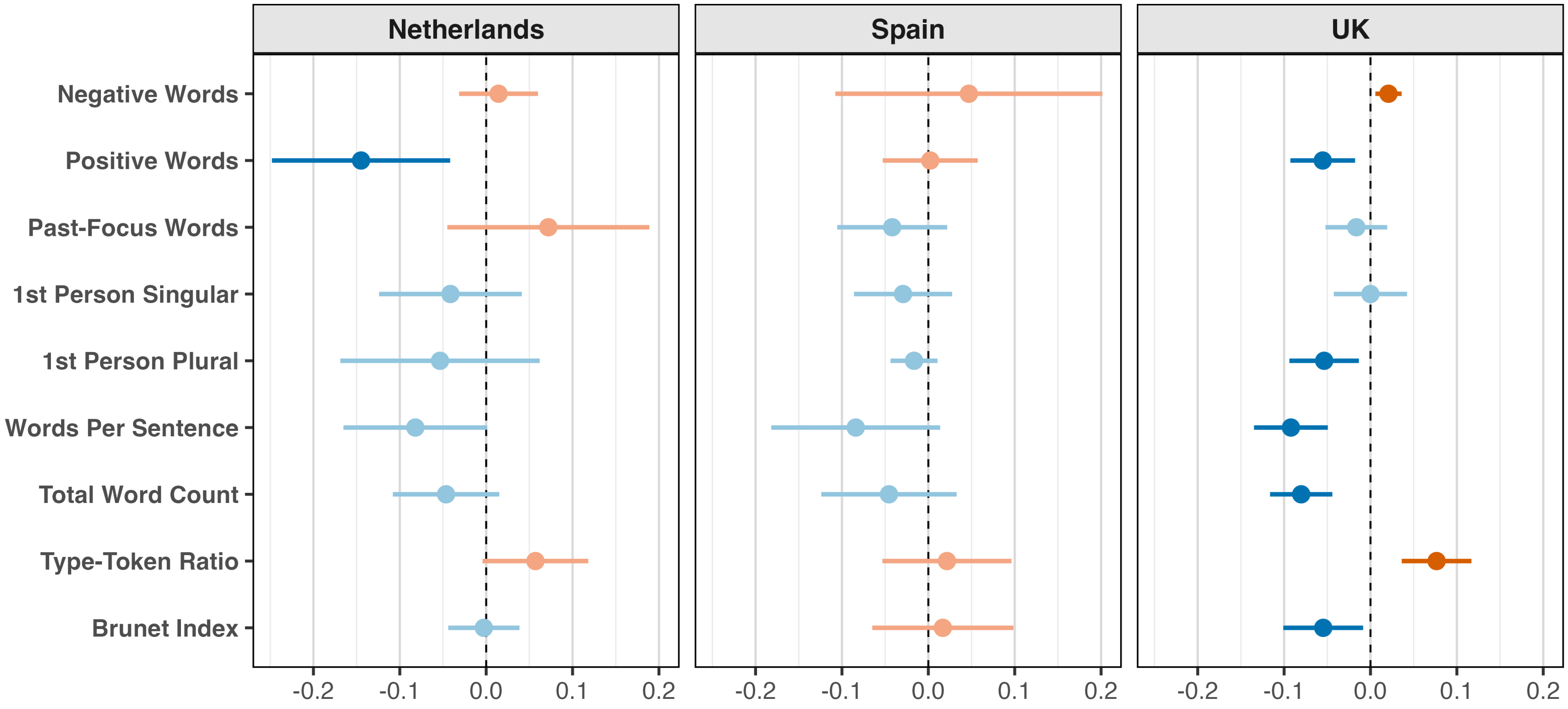